\documentclass[10pt,twocolumn,letterpaper]{article}

\usepackage[pagenumbers]{cvpr} 

\usepackage{graphicx}
\usepackage{amsmath}
\usepackage{amssymb}
\usepackage{booktabs}
\usepackage{pifont}
\usepackage{algpseudocode}
\usepackage{algorithmicx,algorithm}
\usepackage{multicol}
\usepackage{lipsum}
\usepackage{mwe}
\usepackage[pagebackref,breaklinks,colorlinks]{hyperref}

\usepackage[capitalize]{cleveref}
\crefname{section}{Sec.}{Secs.}
\Crefname{section}{Section}{Sections}
\Crefname{table}{Table}{Tables}
\crefname{table}{Tab.}{Tabs.}

\def\cvprPaperID{4294} 
\def\confName{CVPR}
\def\confYear{2022}

\begin{document}

\title{Learning of
Global Objective for Network Flow in Multi-Object Tracking}


\author{
Shuai Li$^1$ \hspace{5mm} Yu Kong$^1$ \hspace{5mm} Hamid Rezatofighi$^{2}$\vspace{2mm}\\
$^1$Rochester Institute of Technology \hspace{5mm} $^2$Monash University\\
{\tt\small \{sl6009, Yu.Kong\}@rit.edu} \quad \tt\small {hamid.rezatofighi@monash.edu}
\vspace{3mm}
}
\maketitle

\begin{abstract}
This paper concerns the problem of multi-object tracking based on the min-cost flow (MCF) formulation, which is conventionally studied as an instance of linear program. Given its computationally tractable inference, the success of MCF tracking largely relies on the learned cost function of underlying linear program. Most previous studies focus on learning the cost function by only taking into account two frames during training, therefore the learned cost function is sub-optimal for MCF where a multi-frame data association must be considered during inference. 
In order to address this problem, in this paper we propose a novel differentiable framework that ties training and inference together during learning by solving a bi-level optimization problem, where the lower-level solves a linear program and the upper-level contains a loss function that incorporates global tracking result. By back-propagating the loss through differentiable layers via gradient descent, the globally parameterized cost function is explicitly learned and regularized. With this approach, we are able to learn a better objective for global MCF tracking. As a result, we achieve competitive performances compared to the current state-of-the-art methods on the popular multi-object tracking benchmarks such as MOT16, MOT17 and MOT20.\vspace{-15pt}
\end{abstract}

\section{Introduction}
\label{sec:intro}

While being a classical problem, multi-object tracking (MOT)~\cite{milan2016mot16, sun2020survey} has been yet one of the most active research areas in computer vision, as being a fundamental basic-level perception task for many real-world problems, \eg, in visual surveillance, and autonomous driving~\cite{geiger2012we}. Thanks to the great progress in object
detection~\cite{felzenszwalb2009object,ren2015faster} techniques, ``tracking-by-detection'' paradigm has dominated the tracking community recently. Given an input video, a set of detection hypotheses is first generated for each frame and the goal of tracking is to associate these detection responses across time, locally or globally, to form all the trajectories. Among various previous works, minimum-cost network-flow~\cite{zhang2008global,pirsiavash2011globally,berclaz2011multiple} based methods have gained increasingly attention due to its fast inference property. In this work, we specifically focus on the network-flow based tracking.

The min-cost network flow formulation for multi-object tracking problem is, indeed, an instance of constrained integer linear program (ILP) with uni-modular constraint matrices~\cite{zhang2008global}. Therefore, the solution to such an ILP problem can be optimally obtained by solving its relaxed version, \ie a constrained linear program (LP), which has an identical optimal integer solution to its ILP counterpart~\cite{andriyenko2010globally}. Given its computationally tractable inference, the success of a network-flow based multi-object tracking approach largely depends on designing a proper cost function. 
Many previous works have focused on learning a robust objective function, \eg the matching cost, between detections in two
frames utilizing a neural network trained using a, \eg binary
cross-entropy~\cite{tang2016multi}, triplet~\cite{chen2018real} or contrastive~\cite{leal2016learning} loss. The major drawback of these approaches is that they only consider a limited temporal context during training, \eg, two or three frames. As such, the learned objective function can be sub-optimal as it ignores long-term temporal contexts and associations. Several recent works adopt graph neural networks~\cite{braso2020learning, papakis2020gcnnmatch, dai2021learning} in order to learn a better feature representation for a spatio-temporal graph formulating multi-object tracking problem. However, their training objectives are still constrained locally, \eg, a binary cross-entropy as a local edge loss is employed during the training stage and thus, knowledge about the global tracking result is not yet properly incorporated.

Recently, there have been few attempts to learn a proper objective function for an LP problem representing a global data association in MOT~\cite{schulter2017deep,frossard2018end}. Seminal work of~\cite{schulter2017deep} adds a log-barrier term into the objective function and adopts a change of basis technique to deal with equality constraints in the linear program, as a result, heuristics are involved in choosing the optimal temperature parameters during the interior method's optimization process, and the tracking results are inferior compared to many top-ranked methods in tracking benchmarks~\cite{milan2016mot16,geiger2012we}. The work of~\cite{frossard2018end} performs 3D tracking on video and LiDAR data, followed by a re-projection step to perform 2D tracking, which is computationally demanding. In contrast to their works, we propose a general framework which adopts Bi-level optimization technique embraced with implicit function theorem~\cite{gould2021deep} to perform end-to-end learning of a global cost function for LP based tracking. At the lower-level of our optimization, our framework solves a linear program and the upper-level contains a general loss function that regularizes the tracking solution. By approximating the original linear program as a continuous quadratic program during the forward pass, it is possible to differentiate through the optimal KKT conditions of the relaxed convex quadratic problem. In this way, the cost for data association can be trained end-to-end by back-propagating the gradient of the loss through differentiable layers. In addition, we integrate a stronger observation model~\cite{bergmann2019tracking} compared to~\cite{schulter2017deep}, together with the learned optimal cost function for data association. Our framework achieves competitive results compared to current state-of-the-art approaches on MOT16, MOT17 and MOT20 benchmarks. 
\\
In summary, our main contributions are as follows:
\begin{itemize}
\itemsep -.5em
    \item We adopt the classical min-cost network flow formulation to address multi-object tracking problem and propose a novel bi-level optimization technique which is able to learn a global cost for tracking directly from multi-frame data association results.
    \item In order to address the non-differentiable problem of the constrained linear program, we propose to approximate the original integer linear program as a \textit{continuous} quadratic program and to back-propagate quadratic program solution's gradients \wrt the model's parameters.
    \item The proposed tracking method achieves results comparable to current state-of-the-art trackers on the popular MOT16, MOT17 and MOT20 benchmark, demonstrating its effectiveness. 
\end{itemize}

\section{Related Works}
\paragraph{Multi-Object Tracking} 
Multi-object tracking remains an active field in computer vision for many years. The way of solving multi-object tracking can be roughly divided into two mainstream approaches namely online and offline methods. Online methods, make the decision by the observations up to the current frame. Popular approach of~\cite{wojke2017simple} employs Hungarian Algorithm~\cite{KM} to associate observations to tracked objects first then use Kalman filter to update the object's states in an recursive manner. JPDAF~\cite{rezatofighi2015joint} extends Global Nearest Neighbor matching principle by allowing all observations to track associations within certain gating areas, making the solutions more robust at the cost of heavy computation. MCMC based data association methods \cite{khan2004mcmc,brau2013bayesian} provide a probabilistic formulation of data association and therefore incorporate arbitrary priors. There also exists tracking methods which make use of deep neural networks. Seminal work of Milan \emph{et al.}~\cite{milan2017online} uses LSTM to address state estimation and data association jointly. Later work such as Fang's \emph{et al.}~\cite{fang2018recurrent} tracked each object through RNN in real-time by coupling the internal and external memory cells. Although online methods can be used in time-critical situations, they make non-reversible decisions, due to the greedy fashion in the data association step.

Offline methods~\cite{segal2013latent,hornakova2021making} for MOT usually constructs a graph whose nodes are the detection hypothesis and edges are the potential links between detection hypothesis, by optimizing a well designed objective function with physically plausible constraints, the final tracking solution can be found. Among them, network-flow~\cite{zhang2008global,pirsiavash2011globally,berclaz2011multiple} based approaches have become popular due to its fast inference and global-optimal solution property, while more robust solutions can be achieved by employing higher-order terms~\cite{roshan2012gmcp,choi2015near,tang2017multiple,hornakova2021making} at the cost of heavy computation. 
\vspace{-15pt}

\paragraph{Graph Neural Networks for MOT} 
Multi-object tracking is essentially a graph optimization problem, and there are several works that attempt to solve tracking by adopting Graph Neural Networks~\cite{kipf2016semi,gilmer2017neural}. Early work of~\cite{jiang2019graph} combines CNN and LSTM to learn appearance and motion features together followed by GCN~\cite{kipf2016semi} for feature refinement. Li \etal~\cite{GraphNetworks} designs an appearance and motion graph network separately for feature learning using a modified message passing network, for online tracking. The recent work from Dai \etal~\cite{dai2021learning} clusters and ranks tracklets through a graph convolution network and shows promising results. Braso and Leal-Taxie~\cite{braso2020learning} leverages message passing networks for network flow based tracking but their work optimizes a binary cross-entropy loss during training and does not allow learning from data association directly, also a heuristic rounding step is applied to ensure disjoint path constraints. By contrast, our method directly perform back-propagation from data association and do not require heuristics at inference stage.
\vspace{-15pt}

\begin{figure*}[!bht]
\centering
\begin{subfigure}{0.5\textwidth}
  \centering
  \includegraphics[width=1.0\linewidth]{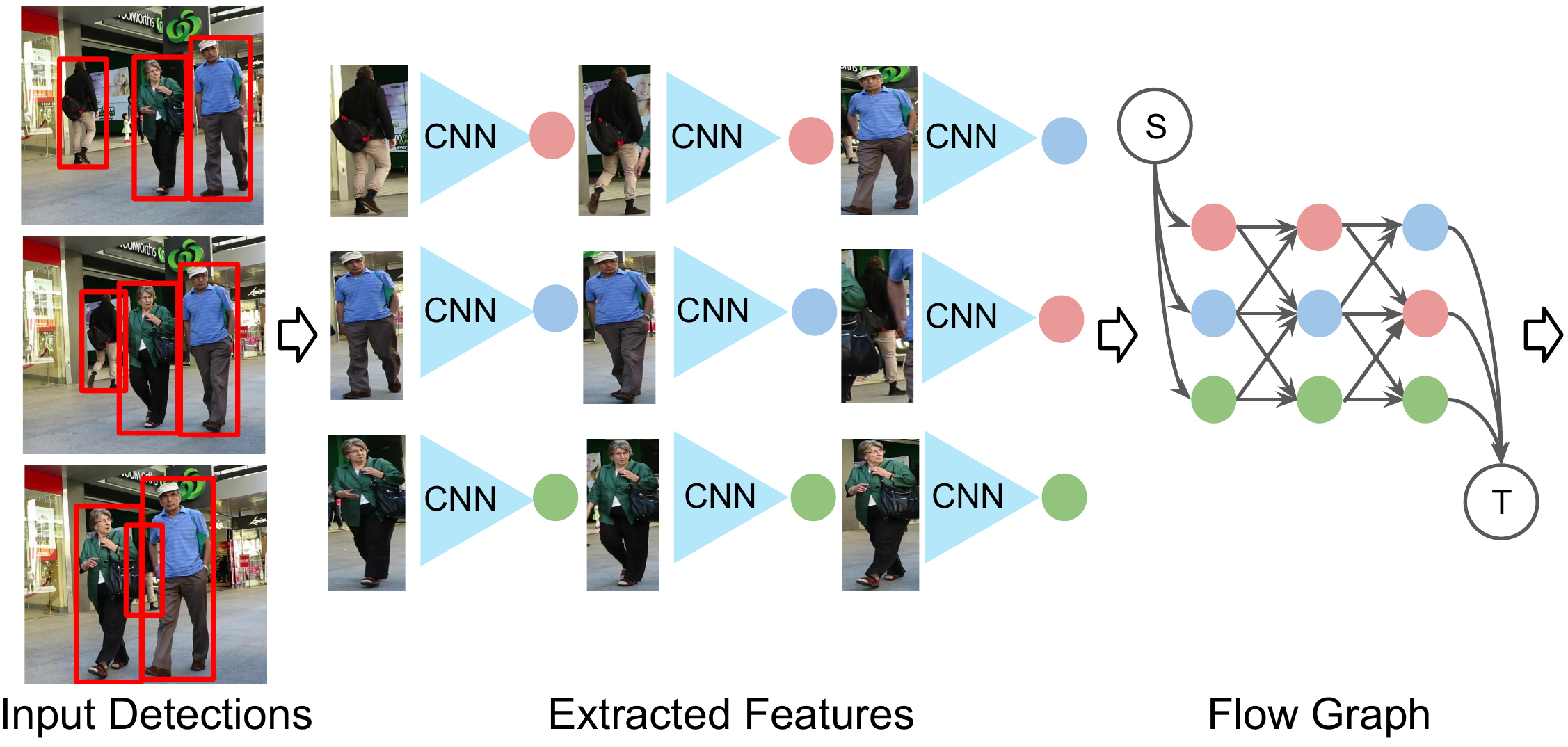}
  \label{fig:sub1}
\end{subfigure}%
\begin{subfigure}{0.5\textwidth}
  \centering
  \includegraphics[width=1.0\linewidth]{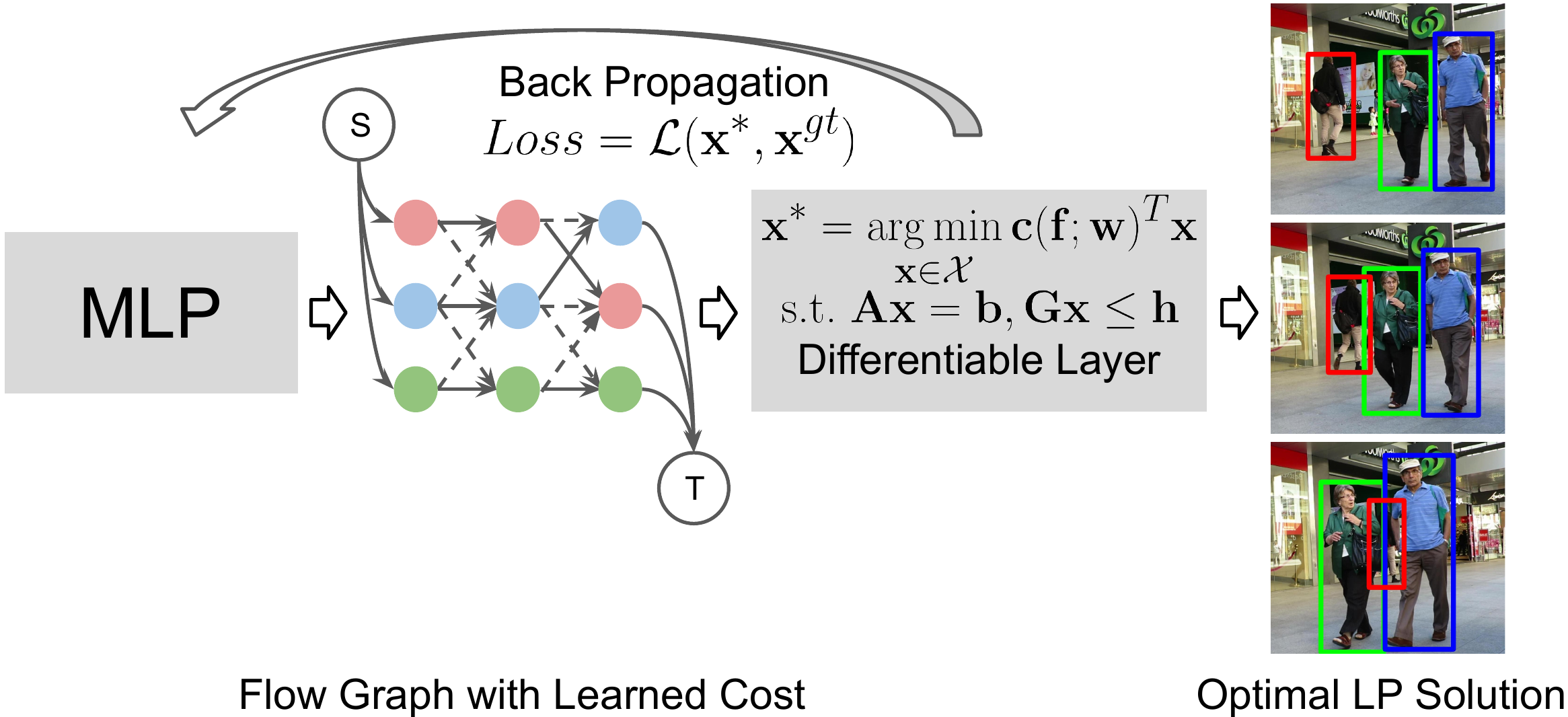}
  \label{fig:sub2}
\end{subfigure}
\caption{Illustration of the proposed tracking method. Given a sequence of frames and a set of detection hypotheses as input. Detection's appearance features extracted by a pretrained person Re-Identification network are combined with geometric cues to build a directed flow graph where detections represent nodes and edges connect detections across frames. A MLP is used to regress the linking probability between detections. During training the lower-level linear program generates a prediction $\mathbf{x}^\ast$, which is passed through a differentiable layer to produce a loss $\mathcal{L}$ from upper-level, the loss is back-propagated through previous layers in order to learn an optimal parameterized cost $\mathbf{c}(\mathbf{f};\mathbf{w})$. At inference time the model outputs data assocition by solving a linear program and achieves tracking. $\mathbf{A,b,G,h}$ denote the Linear Program's constraints.}
\label{fig:pipeline}
\end{figure*}

\paragraph{End-to-End Learning in MOT}
There exists several works that attempt to learn affinity measure for tracking in an end-to-end fashion. The framework in~\cite{xu2020train} utilizes a GRU to approximately differentiate Hungarian Algorithm and achieves descent performance. Burke and Ramamoorthy~\cite{burke2021learning} adopts Sinkhorn Network within Kalman Filtering framework to learn association costs using an EM algorithm, but can only track fixed objects. Similar to our work, Papakis \etal~\cite{papakis2020gcnnmatch} proposes a differentiable matching layer using Sinkhorn Network, while our work is a generalization of their work in multi-frame case. Peng~\emph{et~al.}~\cite{peng2020chained} perform joint detection and data association using a deep CNN. He \etal~\cite{he2021learnable} presents an end-to-end learnable graph matching method but its quadratic formulation largely slows down the inference speed due to its exponential complexity. All of these works perform learning in an online manner. In contrast, our work performs end-to-end learning for global data association from multiple frames and is more robust than online methods during inference.

\section{Approach}

In this section, we briefly revisit the minimum cost network flow formulation for solving multi-object tracking problem, then present our proposed end-to-end learning strategy which performs back-propagation through optimal linear program solutions that learns a suitable cost function for tracking task.

\subsection{Minimum-Cost Network Flow Problem}
Given a set of detection hypothesis $\mathcal{D}=\{\mathbf{d}_i\}$, where $\mathbf{d}_i=(t_i,x_i,y_i,w_i,h_i,s_i)$ denotes a detection at frame $t_i$ located at position $x_i,y_i$ with bounding box size $w_i,h_i$ and confidence score $s_i$ respectively, the goal of tracking is to seek a set of $K$ trajectories $\mathcal{T}=\{T_k\}$ according to bayes rule, which maximizes the posterior probability of data association given input detections: $P(\mathcal{T}|\mathcal{D}) = \frac{P(\mathcal{D}|\mathcal{T})P(\mathcal{T})}{P(\mathcal{D})}$, where $P(\mathcal{D})$ is a normalizing constant that does not influence the solution. Assuming the tracks are independent with each other, and the detections are conditionally independent given the tracks, we aim to optimize:

\begin{equation}\label{MAP}
\begin{split}
\mathcal{T}^{\ast} =& \mathop{\arg\max}_\mathcal{T} p(\mathcal{T}) \cdot \prod_i p(\mathbf{d}_i|\mathcal{T})\\
=& \mathop{\arg\max}_{\mathcal{T}} \prod_k p(T_k) \cdot \prod_i
p(\mathbf{d}_i|\mathcal{T})\\
\end{split}
\end{equation}
where $p(\mathbf{d}_i|\mathcal{T})$ is the likelihood of observing detection $\mathbf{d}_i$ within a track, a Bernoulli distribution is modeled with the output of pedestrian detection. $p(T_k)$ denotes the probability of choosing a sequence of detections for track $T_k$. A first-order Markovian assumption is placed for a specific track $T_k$, the probability can be factorized as:

\begin{equation}
p(T_k) = p_{en}(\mathbf{d}_1) \left( \prod_{i=1}^{l-1}p_{tran}(\mathbf{d}_{i+1}|\mathbf{d}_i) \right) p_{ex}(\mathbf{d}_l)
\end{equation}

Specifically, $p_{en}(\mathbf{d}_1), p_{ex}(\mathbf{d}_l)$ denotes the trajectory having length $l$, the flow enters at detection $\mathbf{d}_1$ and exits at detection $\mathbf{d}_l$, $p_{tran}(\mathbf{d}_{i+1}|\mathbf{d}_i)$ models the temporal transition prior that detection $\mathbf{d}_{i+1}$ follows $\mathbf{d}_i$ within a certain trajectory. We formulate the above tracking problem as a minimum cost network-flow problem by taking the negative logarithm of Eq~\ref{MAP}, and incorporating the disjoint path constraint as well as flow conservation constraint that flow coming to a node is equal to the flow coming out of a node~\cite{zhang2008global}. As such, the above problem can be converted to a constrained integer linear program:
\begin{equation}\label{eq:LP}
\begin{aligned}
&\mathbf{x}^{\ast} = \mathop{\arg\min}_{\mathbf{x} \in \mathcal{X}} \mathbf{c}(\mathbf{f};\mathbf{w})^T \mathbf{x}\\
&\text{s.t.} \ \mathbf{Ax}=\mathbf{b}, \mathbf{Gx} \leq \mathbf{h}
\end{aligned}
\end{equation}

where $\mathbf{x} \in \{0,1\}^n$ is a binary vector consisting of all the edges in the flow graph. $\mathbf{c}(\mathbf{f};\mathbf{w})$ is the parameterized cost function with $\mathbf{f}$ and $\mathbf{w}$ denote input features and the corresponding classifier, $\mathbf{A} \in \mathbb{R}^{2m \times n} ,\mathbf{b} \in \mathbb{R}^{2m} $ and $\mathbf{G} \in \mathbb{R}^{2m \times n},\mathbf{h} \in \mathbb{R}^{2m}$ denote the equality and inequality constraints of the linear program respectively, where $m$ is the number of detections in the flow graph. Note that although we adopt an interior point method to solve the linear program, other optimization techniques, such as maximum-weighted clique, $K$-Shortest path algorithm can be adopted during inference, given the designed cost function.

\subsection{End-to-End Learning of Cost Functions for Min-Cost Flow}
 Figure~\ref{fig:pipeline} illustrates our proposed training pipeline. At the lower level, we solve a linear program, where the cost
 $\mathbf{c}=[\mathbf{c}^{det},\mathbf{c}^{en},\mathbf{c}^{ex},\mathbf{c}^{tran}]$. Specifically, for a detection $\mathbf{d}_i$, $\mathbf{c}^{det}_{i}$ is the detection cost, $\mathbf{c}^{en}_{i},\mathbf{c}^{ex}_{i}$ are designed such that a track starts or ends at this detection. $\mathbf{c}^{tran}$ is a vector consisting of transition costs between two detections. The loss $\mathcal{L}$ at upper-level characterizes the difference between the solution produced by the LP during forward pass and the corresponding ground truth. In order to learn the parameterized cost function $\mathbf{c}(\mathbf{f;w})$ in Eq.~\ref{eq:LP}, we need to calculate the gradient of loss w.r.t. $\mathbf{w}$: $\frac{d{\mathcal{L}}}{d{\mathbf{w}}} = \frac{d{\mathcal{L}}}{d{\mathbf{x}^\ast}}\frac{d{\mathbf{x}^\ast}}{d{\mathbf{c}}}\frac{d{\mathbf{c}}}{d{\mathbf{w}}}$. Therefore, our formulation requires differentiating through the optimal linear program's solution during training. $\frac{d{\mathcal{L}}}{d{\mathbf{x}}}$ and $\frac{d{\mathbf{c}}}{d{\mathbf{w}}}$ are easy to calculate, while computing $\frac{d{\mathbf{x}^\ast}}{d{\mathbf{c}}}$ is difficult, which requires differentiating through an $\mathop{\arg \min}$ operator. Furthermore, the solutions of the linear program are inherently discrete and need to satisfy certain constraints which further complicates the problem.
 
 \noindent{\textbf{Back-propagation through Linear Program}}
 Inspired by the work of Amos~\cite{amos2017optnet}, we propose to back-propagate through linear program's solution at the optimal Karush–Kuhn–Tucker (KKT) condition, and explicitly adopt the implicit function theorem~\cite{barratt2018differentiability,gould2021deep} at this condition. Specifically, for our optimization problem in Eq.~\ref{eq:LP}, its Lagrangian is given by:
 
\begin{equation}\label{eq:Lagrangian}
L(\mathbf{x}, \boldsymbol\lambda, \boldsymbol\nu) = 
f(\mathbf{x}) + \boldsymbol \lambda^T (\mathbf{Gx} - \mathbf{h}) + \boldsymbol\nu^T (\mathbf{Ax} - \mathbf{b})
\end{equation}

where $f(\mathbf{x}) = \mathbf{c}^T \mathbf{x}$ is the linear objective, $\boldsymbol\lambda \geq 0$ and $\boldsymbol\nu$ are the dual variables which correspond to inequality and equality constraints, respectively. Taking into account stationarity condition, complementary slackness and primal feasibility of the convex problem's KKT condition and apply implicit function theorem on top of these equations, we can get the following matrix equation~\cite{barratt2018differentiability}:

\begin{equation}\label{eq:KKT}
\setlength{\arraycolsep}{1pt}
\begin{bmatrix}
\nabla_{\mathbf{x}}^2f(\mathbf{x}) & \mathbf{G}^T  & \mathbf{A}^T \\
diag(\boldsymbol\lambda)\mathbf{G}  & diag(\mathbf{G}\mathbf{x}-\mathbf{h}) & \mathbf{0} \\
\mathbf{A} & \mathbf{0}  & \mathbf{0} 
\end{bmatrix}
\begin{bmatrix}
\frac{d\mathbf{x}}{d\mathbf{c}} \\ \frac{d\boldsymbol\lambda}{d\mathbf{c}} \\ \frac{d\boldsymbol\nu}{d\mathbf{c}} 
\end{bmatrix}
=
\begin{bmatrix}
 -\frac{d\nabla_\mathbf{x} L(\mathbf{x},\boldsymbol \lambda, \boldsymbol \nu)}{d\mathbf{c}} \\ \mathbf{0} \\ \mathbf{0}
\end{bmatrix}
\end{equation}

Here, $diag(\cdot)$ operation converts a vector into a diagonal matrix. By solving Eq.~\ref{eq:KKT}, the desired Jacobian
$\frac{d{\mathbf{x}}}{d{\mathbf{c}}}$ can be acquired. However, directly doing so is not feasible, since $\nabla_{\mathbf{x}}^2f(\mathbf{x})$ would become $\mathbf{0}$ due to the linear objective. As a result, the left-hand side of Eq.~\ref{eq:KKT} would become singular, and
$\frac{d{\mathbf{x}}}{d{\mathbf{c}}}$ would be trivials.

In order to tackle this problem, we propose to add a Tikhonov damping term $\gamma$ into the original linear objective so that $f(\mathbf{x}) = \mathbf{c}^T \mathbf{x} + \gamma ||\mathbf{x}||_{2}^2$, and further relax the $\mathbf{x}$ such that $\mathbf{x} \in [0,1]$ to enable gradient-based optimization. Therefore, the original linear program in Eq.~\ref{eq:LP} essentially becomes a \textit{continuous} quadratic program (QP):

\begin{equation}\label{eq:QP}
\begin{aligned}
\mathbf{\hat{x}} =& \mathop{\arg\min}_{\mathbf{x} \in \mathcal{X}} \frac{1}{2}\mathbf{x}^T \mathbf{Qx} + \mathbf{c}^T \mathbf{x} \\
&\text{s.t.} \ \mathbf{Ax}=\mathbf{b}, \mathbf{Gx} \leq \mathbf{h}
\end{aligned}
\end{equation}

In particular, $\mathbf{Q} \in \mathbb{R}^{n \times n}$, and $\mathbf{Q} \succ \mathbf{0}$, so the quadratic objective is strictly convex. Since $\nabla_{\mathbf{x}}^2f(\mathbf{x}) = \mathbf{Q}$, the new linear system can be written as Eq.~\ref{eq:KKT1}. By solving this KKT equation during the forward pass, we can get the desired Jacobian $\frac{d\mathbf{\hat{x}}}{d\mathbf{c}}$ and back-propagate the QP in order to perform gradient-based end-to-end training.

\begin{equation}\label{eq:KKT1}
\setlength{\arraycolsep}{1pt}
\begin{bmatrix}
\mathbf{Q} & \mathbf{G}^T  & \mathbf{A}^T \\
diag(\boldsymbol\lambda)\mathbf{G} & diag(\mathbf{G}\mathbf{x}-\mathbf{h}) & \mathbf{0} \\
\mathbf{A} & \mathbf{0}  & \mathbf{0} 
\end{bmatrix}
\begin{bmatrix}
\frac{d\mathbf{x}}{d\mathbf{c}} \\ \frac{d\boldsymbol\lambda}{d\mathbf{c}} \\ \frac{d\boldsymbol\nu}{d\mathbf{c}} 
\end{bmatrix}
=
\begin{bmatrix}
 -\frac{d\nabla_\mathbf{x} L(\mathbf{x},\boldsymbol \lambda, \boldsymbol \nu)}{d\mathbf{c}} \\ \mathbf{0} \\ \mathbf{0}
\end{bmatrix}
\end{equation}

As for the lower-level's loss, we employ $L_2$ loss, which directly measures the difference between the predicted data association $\mathbf{\hat{x}}$ and the ground truth assignment $\mathbf{x}^{gt}$ as $||\mathbf{\hat{x}}-\mathbf{x}^{gt}||_{2}^2$. Note that the other loss functions between two binary vector, \eg, hamming loss, can also be used here. Our experiment shows that our framework is not very sensitive to the tuned value of $\gamma$ as long as $\gamma$ is small. Therefore, we set $\gamma=0.1$. The full training algorithm is detailed in Algorithm~\ref{alg1}. 

\begin{algorithm}
\caption{Gradient descent for end-to-end learning of cost function for network flow.}
\label{alg1}
\hspace*{0.02in} {\bf Input:} 
Training set $\mathcal{D}_{train} = \{(\mathbf{f}_i,\mathbf{x}_i^{gt})\}_{i=1}^N$ with $N$ flow graphs, each paired with feature representation $\mathbf{f}_i$ and ground truth data association $\mathbf{x}_i^{gt}$\\
\hspace*{0.02in} {\bf Output:}
MLP with learned optimal model parameters $\mathbf{w^\ast}$
\begin{algorithmic}[1]
\State {Initialize learning rate $\alpha$, MLP with $\mathbf{w}_{0}$} 
\Repeat
\State Randomly sample $M$ graphs from $\mathcal{D}_{train}$.
\For{$i=1$ to $M$}
\State Forward $\mathbf{f}_i$ to MLP to obtain cost $\mathbf{c}_i$, and solve Eq.~\ref{eq:QP} to get $\hat{\mathbf{x}}_i$, calculate loss: $\mathcal{L}(\hat{\mathbf{x}}_i, \mathbf{x}_i^{gt})$
\State Calculate gradient of $\mathcal{L}$: $\frac{d\mathcal{L}}{d\mathbf{w}_i}$ =
$\frac{d\mathcal{L}}{d\hat{\mathbf{x}}_i}\frac{d\hat{\mathbf{x}}_i}{d\mathbf{c}_i}\frac{d\mathbf{c}_i}{d\mathbf{w}_i}$.
\EndFor
\State Set $\frac{d\mathcal{L}}{d\mathbf{w}} = \frac{1}{M}\sum_{i=1}^M \frac{d\mathcal{L}}{d\mathbf{w}_i}$, perform gradient descent step: $\mathbf{w}_{t+1}$ = $\mathbf{w}_t$ - $\alpha \frac{d \mathcal{L}}{{d \mathbf{w}}}$
\Until{Convergence}
\end{algorithmic}
{\bf Return} Learned MLP parameterized by $\mathbf{w^\ast}$
\end{algorithm}

\subsection{Network Flow Cost Function}
Considering trained MLPs with parameters $\mathbf{w}^\ast$, it is possible to design the unary and binary potentials defined in the flow graph for inference.

\noindent{\textbf{Detection Cost.}}
Given a detection $\mathbf{d}_i$, the unary cost $\mathbf{c}_i^{det}$ is defined as $-s_i$, where $s_i$ is the detection confidence output by a class-specific classifier. This term favors high-confidence person detections to be selected in the tracking results.

\noindent{\textbf{Entry/Exit Cost.}}
These costs are learned scalars such that during linear program's inference, a longer-track is more likely to be selected. We cross-validated on the training set and found that a scalar 1 works well in practice. Note that a high entry/exit cost trivially yields all-zero solution of the LP as it increases total cost/energy.

\noindent{\textbf{Transition Cost.}}
Given a pair of detections $\mathbf{d}_{i}^t$ and $\mathbf{d}_{j}^{t+1}$ and their edge feature representation $\mathbf{e}_{ij}$, the matching probability is: $p_{ij} = \text{MLP}(\mathbf{e}_{ij};\mathbf{w}^\ast)$. After that, $\mathbf{c}_{ij}^{tran}$ is set to $-\log{p_{ij}}$, which suggests detection pairs with high matching probabilities should be connected.

\vspace{-0.2cm}
\section{Experiments}
\subsection{Datasets}
We conduct the experiments on MOT16, MOT17~\cite{milan2016mot16} and MOT20~\cite{dendorfer2020mot20} pedestrian tracking dataset. MOT16 and MOT17 contain the same videos, except that MOT16 applies DPM~\cite{felzenszwalb2009object} detections as input for tracking, while MOT17 evaluates tracking performance thoroughly under three different detection inputs, namely DPM~\cite{felzenszwalb2009object}, FRCNN~\cite{ren2015faster} and SDP~\cite{yang2016exploit}. Further, MOT20 has been designed to challenge the tracking algorithm's ability to track crowded scenes~\cite{dendorfer2020mot20}. 

Since MOT16 and MOT17 has almost similar ground truth annotation, we train our model on the MOT16 training set. We use MOT16-09, MOT16-13 sequence to form our validation set and the remaining 5 sequences as the training set. For fair comparison with other methods, the tracking performance is reported in MOT16, MOT17 and MOT20 test set using the provided public detections.

\subsection{Implementation Details}
\noindent{\textbf{Detections.}}
As network-flow based methods are very sensitive to false positives. We first pre-process the raw input detections using~\cite{bergmann2019tracking} provided detections as suggested by~\cite{hornakova2020lifted}, in order to obtain a set of high quality detections.

\noindent{\textbf{Features.}}
Given a pair of detections $\mathbf{d}_i$ and $\mathbf{d}_j$ with $(t_i,x_i,y_i,w_i,h_i,s_i)$ and $(t_j,x_j,y_j,w_j,h_j,s_j)$ respectively, we follow the work of~\cite{braso2020learning} to encode spatial-temporal constraint as the geometric feature: $(\frac{2(x_j-x_i)}{h_i+h_j},\frac{2(y_j-y_i)}{h_i+h_j},\log\frac{h_i}{h_j},\log\frac{w_i}{w_j})$.

This constraint encodes important relative position information between two detections, the intuition is that under a small time gap, the pedestrians cannot move far away, and the size of the pedestrian should not change much due to markovian property.

In addition to the above mentioned spatial-temporal features, we also incorporate appearance features of person detections. For this, we adopt the pre-trained deep ReID~\cite{zhou2019omni} architecture, which is the state-of-the-art model in the person re-identification literature, and use the pre-trained model to extract accurate appearance feature for each detection. The normalized cosine distance between two detection's appearance features output by the ReID network $\phi$ is concatenated with the above mentioned feature together with the generalized intersection over union (GIoU)~\cite{rezatofighi2019generalized} metric, which is more robust to geometric deformation, as the final edge feature cost regression:

\begin{scriptsize}
\begin{equation}
    \mathbf{e}_{ij}=(\frac{2(x_j-x_i)}{h_i+h_j},\frac{2(y_j-y_i)}{h_i+h_j},\log\frac{h_i}{h_j},\log\frac{w_i}{w_j},{\phi(\mathbf{d}_i)}^T \phi(\mathbf{d}_j), \mathrm{GIoU}(\mathbf{d}_i,\mathbf{d}_j)).
\end{equation}
\end{scriptsize}

\noindent{\textbf{Training.}}
In order to generate training samples, we sub-divide each training sequence equally into (overlapping) $T$ frames, where we set $T$ = 15. Note that we only consider connections of nodes in two adjacent frames to avoid heavy memory consumption and speed up calculation. We utilize the available ground-truth annotations which defines the ideal data association between objects across frames. In particular, for each item in $\mathbf{c}^{tran}$ we set $-\log{p}_{ij}$, where $p_{ij}$ is the matching probability for ground-truth bounding box $\mathbf{d}_i$ and $\mathbf{d}_j$. $-1$ is used for entries in $\mathbf{c}^{det}$ as each annotated box is a true positive. For each ground-truth box $\mathbf{d}_i$, we connect it with the corresponding entry and exit nodes with $\mathbf{c}_i^{en}$ and $\mathbf{c}_i^{ex}$ set to 1 as we make a uniform prior distribution of the start and termination over each nodes. 

We experimented with different network architectures for scoring the cost function, \eg linear classifier, MLP, and found that a two-layer multi-layer perceptron works better in practice. Therefore, we adopt a two layer MLP with ReLU non-linearity which outputs a probability distribution between 0 and 1 as affinity measure, using Adam optimizer with initial learning rate of $10^{-3}$ with a weight decay of $10^{-4}$ for around 10 epochs. We select the model that performs best on the validation set for tracking at test time. 

\noindent{\textbf{Inference.}}
Tracking is performed over (overlapping) batches with length of 50-150 frames depending on the detection density in a specific video. Within each batch, a maximum of $\Delta=5$ frames gap between detections is allowed to join detections, to handle false negatives and short term occlusions. We utilize Gurobi's solver to solve each ILP within each batch to obtain tracks. Tracks are finally stitched across adjacent batches to form the final tracks.

\noindent{\textbf{Long-Term Occlusion Handling.}}
Due to long-term occlusions in reality, tracks are usually fragmented and identity switches occur. In response, we propose to handle these issues using a second round of network-flow tracker, except that the nodes now are tracklets (short tracks) instead of nodes in the first round, with appearance and motion constraints. Specifically, given tracklet $T_i$, we use~\cite{zhou2019omni} to extract appearance features, \eg $\mathbf{f}_i^{app} = \frac{1}{L} \sum_{d=1}^L \mathbf{f}_d^{app}$ denotes its final representation. Therefore for track $T_i$ and $T_j$, $1 - {\mathbf{f}_i^{app}}^T \mathbf{f}_j^{app}$ is their final matching cost. Regarding motion constraint, we estimate the average velocity for each pair of temporally non-overlapping tracklet, such that $\tau_{dist}$ is used to reject physically implausible connections. As such, tracklets are joined and long-term occlusions within a track could be recovered throughs bi-linear interpolation. 

\begin{figure}
\begin{multicols}{3}[\columnsep=0.0cm]
\centering
\includegraphics[scale=0.25]{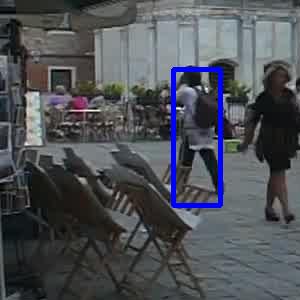}
\columnbreak
\centering
\includegraphics[scale=0.25]{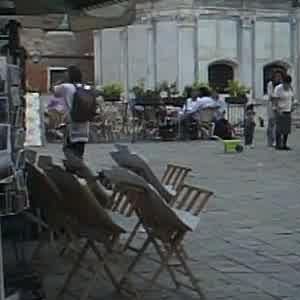}
\columnbreak
\centering
\includegraphics[scale=0.25]{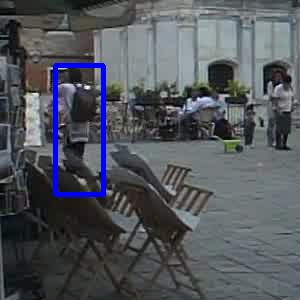}
\end{multicols}
\vspace{-1em}
\captionof{figure}{Effect of applying single-object tracker. Left: Target to track. Middle: Tracker fails to track target due to missing detection. Right: The SOT tracker is able to follow target for a few frames.}\vspace{-1em}
\label{fig:SOT}
\end{figure}

\noindent{\textbf{Post-Processing.}}
It is possible that our tracker lost several objects during tracking, due to the drastic appearance change/illumination, etc. In order to address this problem, we add a single object tracker (SOT) to keep track of the lost objects. Specifically, for a track that disappears before the video reaches to the last frame, we make use of the last position of the tracker to initialize the single object tracker~\cite{lukezic2017discriminative} and perform tracking. In order to kill the tracker in case the tracked object confuses with other objects, we compare the tracked object's appearance with the initialized object, whenever their similarity of color histogram fall below a threshold $\tau$, the tracker is terminated to avoid tracking false positives or drift to background. As can be seen in Figure~\ref{fig:SOT}, by adding a SOT, our method successfully tracks the lost object, and the SOT tracker ends when a complete occlusion occurs.

\subsection{Evaluation}
In order to validate the effectiveness of the proposed end-to-end learning method, we compare our learned MLP against a baseline method, which is trained with binary cross-entropy (BCE) objective using annotated data association, which does not allow back-propagation from association. $L_1$ and $L_2$ are the proposed end-to-end learning approaches which adopt $L_1$ and $L_2$ as the loss, respectively. All methods takes the same edge features as input for a fair comparison.

\noindent{\textbf{Evaluation of Tracking.}}
Multiple Object Tracking Accuracy (MOTA)~\cite{bernardin2008evaluating} and IDF1~\cite{ristani2016performance} are the two widely adopted metrics for tracking performance evaluation. The two metrics are defined as: MOTA$=1 - \frac{\sum_t \text{FP}_{t}+\text{FN}_{t}+\text{IDS}_{t}}{\sum_t \text{GT}_{t}}$ and IDF1$= \frac{\text{2IDTP}}{\text{2IDTP}+\text{IDFP}+\text{IDFN}}$. MOTA measures the tracking error in terms of number of false positive (FP), false negative (FN) and identity switches (IDS), it focus more on the detection's perspective, while IDF1~\cite{ristani2016performance} reflects the tracker's ability to maintain identities over time and thus concentrates more on tracking side. In addition, Multiple Object Tracking Precision (MOTP), the percentage of Mostly Tracked Targets (ML) and the percentage of Mostly Lost Targets (ML) as well as the number of total fragmented times occur in a trajectory (Frag) constitutes the main evaluation metrics for tracking.

\noindent{\textbf{Effectiveness of the Learned Affinity Measures.}}
Area Under Curve (AUC) is suitable to measure a binary classifier's performance, therefore a stronger MLP should obtain a higher AUC score. Mean Squared Error (MSE) is adopted to compare the QP's output $\mathbf{x}^\ast$, including the detection, entry/exit cost, against ground-truth annotation $\mathbf{x}^{gt}$. MSE Edge solely takes into account the predicted data association compared with ground truth association. From Table~\ref{tab:Affinity}, it is clear that by back-propagating through multi-frame data association, our proposed approach, regardless of the losses used, outperforms baseline method trained using BCE objective in all metrics used. Overall, $L_1$ and $L_2$ loss achieve similar performance.

\vspace{-0.2cm}
\begin{table}[h]
\center
\tabcolsep=0.5cm
\resizebox{\columnwidth}{!}
{\begin{tabular}{l c c c c}
\toprule
     Loss & AUC$\uparrow$ & BCE$\downarrow$ & MSE$\downarrow$ & MSE Edge$\downarrow$\\
\midrule
     BCE & 0.996 & 0.064 & 0.026 & 0.017\\ 
\midrule
    $L_1$ & \textbf{0.997} & 0.047 & 0.013 & 0.008\\ 
    $L_2$ & \textbf{0.997} & \textbf{0.005} & \textbf{0.010} & \textbf{0.006}\\
\bottomrule
\end{tabular}}
\vspace{-0.2cm}
\caption{Evaluation of the affinity metrics and data association results using the proposed training strategy versus baseline methods on the MOT16 validation set.}
\label{tab:Affinity}
\end{table}

\vspace{-0.6cm}
\begin{table}[h]
\center
\small
\tabcolsep=0.12cm
\begin{tabular}{l c c c c c} 
 \hline
  Method & MOTA $\uparrow$ & IDF1 $\uparrow$ & IDS $\downarrow$ & MT $\uparrow$ & ML $\downarrow$ \\
 \hline
 w/o 2$^{nd}$-round MCF & 38.9 & 43.5 & 134 & 22 & 57\\ 
 w 2$^{nd}$-round MCF & \textbf{42.9} & \textbf{55.1} & \textbf{73} & \textbf{28} & \textbf{51} \\
 \hline
\end{tabular}
\caption{Ablation study on MOT16 validation set, first-round MCF achieves descent performance, adding second-round MCF further boosts the performance due to long-term occlusion handling.}
\label{tab:Ablation}
\vspace{-1.5em}
\end{table}

\vspace{-0.2cm}
\begin{table}[h]
\center
\small
\tabcolsep=0.13cm
{\begin{tabular}{l c c c c c c c}
\toprule
     Method & MOTA$\uparrow$ & IDF1$\uparrow$ & MT$\uparrow$ & ML$\downarrow$ 
     & FP$\downarrow$ & FN$\downarrow$ & IDS$\downarrow$\\
\midrule
    Baseline & 49.33 & 58.91 & 100 & 163 & 822 & 24814 & \textbf{157}\\
    $L_1$ & 51.45 & \textbf{60.15} & 123 & 151 & \textbf{810} & 23715 & 186\\
    $L_2$ & \textbf{51.54} & \textbf{60.15} & \textbf{124} & \textbf{147} & 831 & \textbf{23641} & 193\\
\bottomrule
\end{tabular}}
\vspace{-0.2cm}
\caption{Evaluation of tracking performance under different approaches on MOT17 validation set, End-to-end learned cost performs better than baseline trained using BCE loss.}
\label{tab:MOT17Val}
\end{table}

\vspace{-0.7em}
\begin{table}[h]
\center
\tabcolsep=0.1cm
\resizebox{\columnwidth}{!}
{\begin{tabular}{l c c c c c c}
\toprule
     Method & MOTA$\uparrow$ & REC$\uparrow$ & PREC$\uparrow$ & MT$\uparrow$ & IDS$\downarrow$ & FRAG$\downarrow$\\
\midrule
    DNF~\cite{schulter2017deep}(Linear) & 28.25 & 38.01 & 80.09 & 9.67 & 342 & 1620\\
    DNF~\cite{schulter2017deep}(MLP) & 31.10 & 37.53 & 85.88 & 8.51 & 289 & 1562\\
    \midrule
    Proposed Method & \textbf{44.27} & \textbf{45.03} & \textbf{98.84} & \textbf{14.9} & \textbf{260} & \textbf{365}\\
\bottomrule
\end{tabular}}
\vspace{-0.2cm}
\caption{Evaluation of tracking performance under different training strategy on MOT16 validation set, $\uparrow$ means high number is better, $\downarrow$ is opposite. Best performance under each metric is shown in bold-faced font.}\vspace{-1em}
\label{tab:MOT16Val}
\end{table}

\noindent{\textbf{Effectiveness of Second Stage Network Flow Data Association.}} In order to outline the contribution of the first-stage data association on the final tracking performance, we test the proposed method on MOT16 validation set. According to the results in Table~\ref{tab:Ablation}, the first stage NF data association provides decent results (as a good tracklet initialization) for the second round of network flow tracker, where long-term occlusions must be handled. We can also see the performance improvements in second stage network flow tracking compared to solely having the first stage tracking. In-terms of MOTA, the improvements are not as significant as IDF1/IDS, which properly reflects the contribution of the second stage to the long-term occlusions. Therefore the core method (\ie the first stage) plays a crucial rule to the final performance.

\noindent{\textbf{Effectiveness of The Proposed Approach over Baseline in Tracking.}} In Table~\ref{tab:MOT17Val}, we compare our method against a NF baseline trained using BCE objective, in terms of tracking metrics. Note that for all three methods, we use the same strategy of second-round MCF for long-term occlusion handling, thus the difference of final tracking performance merely stems from the first-stage's association results. Though our method is slightly inferior in IDS metric, our proposed approach surpass the baseline in MOTA, and ML metrics, which suggests back-propagation from data association is indeed necessary to reach a stronger tracking performance.

\noindent{\textbf{Comparison against ~\cite{schulter2017deep}.}} We compare our proposed learning method against the approach proposed by Schulter \etal~\cite{schulter2017deep} in Table~\ref{tab:MOT16Val}. We followed the same train/val split as their work and cross-validated in the MOT16 training set for fair comparison. For all tracking metrics, we outperform their end-to-end learning method. The higher recall and precision can be attributed to the fact that we have a stronger observation model, as well as our better learned affinity measure for data association. We improve over the baseline MOTA by $44\%$ and significantly increase the MT/FRAG metric.

\begin{table}[!htbp]
    \centering
     \centering
    \tabcolsep=0.08cm
    \scalebox{0.7}{
    \begin{tabular}{l c c c c c c c c}
    \toprule
    Method & Mode  & MOTA $\uparrow$ & IDF1 $\uparrow$ & MT $\uparrow$ &  ML $\downarrow$ & FP $\downarrow$ & FN $\downarrow$ & IDS $\downarrow$\\
    \midrule
    EAMTT~\cite{sanchez2016online} & Online & 38.8 & 42.4 & 7.9 & 49.1 & 8114 & 102452 & 965 \\
    RAN~\cite{fang2018recurrent} & Online & 45.9 & 48.8 & 13.2 & 41.9 & 6871 & 91713 & 648\\
    AMIR~\cite{sadeghian2017tracking} & Online & 47.2 & 46.3 & 14.0 & 41.6 & \textbf{2681} & 92856 & 774\\
    MOTDT~\cite{chen2018real} & Online & 47.6 & 50.9 & 15.2 & 38.3 & 9253 & 85431  & 792\\
    GraphNetwork~\cite{GraphNetworks} & Online & 47.7 & 43.2 & 16.1 & 34.3 & 9518 & 83875 & 1907\\
    UMA~\cite{yin2020unified} & Online & 50.5 & 52.8 & 17.8 & \textbf{33.7} & 7587 & 81924 & 685 \\
    Tracktor++~\cite{bergmann2019tracking} & Online & 54.4 & 52.5 & 19.0 & 36.9 & \underline{3280} & 79149 & 682\\
     \midrule
    LINF1~\cite{fagot2016improving} & Offline &41.5 &45.7 &11.6 &51.3 &7896 &99224 &430\\
    BiLSTM~\cite{kim2018multi} & Offline &42.1 &47.8 &14.9 &44.4 &11637 &93172 &753\\
    NOMT~\cite{choi2015near} & Offline &46.4 &53.3 &18.3 &41.4 &9753 &87565 &\underline{359}\\
    LMP~\cite{tang2017multiple} & Offline &48.8 &51.3 &18.2 &40.1 &6654 &86245 &481\\
    MPNTrack~\cite{braso2020learning} & Offline &\textbf{58.6} &\textbf{61.7} &\textbf{27.3} &\underline{34.0} &4949 &\textbf{70252} &\textbf{354}\\
   \textbf{LPT}(Ours) & Offline & \underline{57.4} & \underline{58.7} & \underline{22.7} & 37.2 &4201 & \underline{73114} & 427\\
    \bottomrule
    \end{tabular}
    }
    \caption{Tracking results on MOT16 test set with DPM \cite{felzenszwalb2009object} detections as input. Bold and underlined numbers indicate the best and the second best performances.}
    \label{tab:motchallenge_mot16}
\end{table}

\begin{table}[!htbp]
    \centering
     \centering
    \tabcolsep=0.08cm
    \scalebox{0.7}{
    \begin{tabular}{l c c c c c c c c}
    \toprule
    Method & Mode  & MOTA $\uparrow$ & IDF1 $\uparrow$ & MT $\uparrow$ &  ML $\downarrow$ & FP $\downarrow$ & FN $\downarrow$ & IDS $\downarrow$\\
    \midrule
    DMAN~\cite{zhu2018online}& Online & 48.2 & 55.7 & 19.3 & 38.3 & 26128 & 263608 & 2194 \\
    MOTDT~\cite{chen2018real}& Online & 50.9 & 52.7 & 17.5 & 35.7 & 24069 & 250768 & 2474\\
    STRN~\cite{xu2019spatial}& Online & 50.9 & 56.5 & 20.1 & 37.0 & 27,532 & 246924 & 2593 \\
     FAMNet~\cite{chu2019famnet}& Online & 52.0 & 48.7 & 19.1 & 33.4 & 14138 & 253616 & 3072 \\
    DeepMOT~\cite{xu2020train}& Online & 53.7 & 53.8 & 19.4 & 36.6 &\underline{11731} & 247447  & 1947\\
    Tracktor++v2~\cite{bergmann2019tracking}& Online & 56.3 & 55.1 & 21.1 & 35.3 & \textbf{8866} & 235449 & 1987\\
    GCNNMatch~\cite{papakis2020gcnnmatch} & Online & 57.3 & 56.3  & 24.2 & \textbf{33.4} & 14100 & 225042 & 1911 \\
     \midrule
     MHT\_DAM~\cite{kim2015multiple}& Offline & 50.7 & 47.2 & 20.8 & 36.9 & 22875 & 252889 & 2314\\
     jCC~\cite{keuper2018motion}& Offline & 51.2 & 54.5 & 20.9 & 37.0 & 25937 & 247822 & 1802\\
     JBNOT~\cite{henschel2019multiple}& Offline & 52.6 & 50.8 & 19.7 & 35.8 & 31572 & 232659 & 3050\\
     MPNTrack~\cite{braso2020learning} & Offline & \underline{58.8} & \underline{61.7} & \textbf{28.8} & 33.5 & 17413 & \underline{213594} & \textbf{1185}\\
     Lif\_T~\cite{hornakova2020lifted} & Offline & \textbf{60.5} & \textbf{65.6} & \underline{27.0} & \underline{33.6} & 14966 & \textbf{206619} & \underline{1189}\\
    \textbf{LPT}(Ours) & Offline & 57.3 & 57.7 & 23.3 & 36.9 & 15187 & 224560 & 1424\\
    \bottomrule
    \end{tabular}
    }
    \caption{Tracking results on MOT17 test set using public detections \cite{felzenszwalb2009object,ren2015faster,yang2016exploit} as input. The best and second best performances are shown in bold and underlined numbers respectively.}
    \label{tab:motchallenge_mot17}
\end{table}

\begin{table}[!htbp]
    \centering
     \centering
    \tabcolsep=0.08cm
    \scalebox{0.7}{
    \begin{tabular}{l c c c c c c c c}
    \toprule
    Method & Mode & MOTA $\uparrow$ & IDF1 $\uparrow$ & MT $\uparrow$ &  ML $\downarrow$ & FP $\downarrow$ & FN $\downarrow$ & IDS $\downarrow$\\
    \midrule
    SORT20~\cite{bewley2016simple}& Online & 42.7 & 45.1 & 16.7 & 26.2 & 27521 & 264694 & 4470 \\
    Tracktor++V2~\cite{bergmann2019tracking}& Online & 52.6 & 52.7 & 29.4 & 26.7 & \textbf{6930} & 236680 & \underline{1648} \\
    ArTist~\cite{saleh2021probabilistic}& Online & 53.6 & 51.0 & 31.6 & 28.1 & \underline{7765} & 230576 & \textbf{1531}\\
     \midrule
     GCNNMatch~\cite{papakis2020gcnnmatch}& Offline & 54.5 & 49.0 & 32.8 & 25.5 & 9522 & 223611 & 2038\\
     ApLift~\cite{hornakova2021making}& Offline & \textbf{58.9} & \textbf{56.5} & \textbf{41.3} & \textbf{21.3} & 17739 & \textbf{192736} & 2241\\
    \textbf{LPT}(Ours) & Offline & \underline{57.9} & \underline{53.5} & \underline{39.0} & \underline{22.8} & 9980 & \underline{205949} & 1827\\
    \bottomrule
    \end{tabular}
    }
    \caption{Tracking results on MOT20 test set using public detections as input. Bold and underlined numbers indicate the best and the second best performances.}
    \label{tab:motchallenge_mot20}
\end{table}

\noindent{\textbf{Comparison with Other State-of-the-arts.}}
We compare our tracker's performance against other tracking algorithms on MOT16, MOT17 and MOT20 benchmark. In our final results, we selected the models trained using $L_2$ loss for benchmark evaluation. We do not apply SOT in our final implementation as the improvements are marginal. 

Table~\ref{tab:motchallenge_mot16},~\ref{tab:motchallenge_mot17} and~\ref{tab:motchallenge_mot20} compares our method, namely Linear Program Tracker (LPT), against existing methods. On MOT16 benchmark, our method has second-best performace in MOTA, IDF1, MT and FN. Note that the work of~\cite{braso2020learning} use Message Passing Networks that learns a better feature representation for temporal connections, while our method does not utilize Graph Networks, and our result is still comparable with their approaches.

\begin{figure*}[hbt!]
\begin{center}
\includegraphics[width=\textwidth]{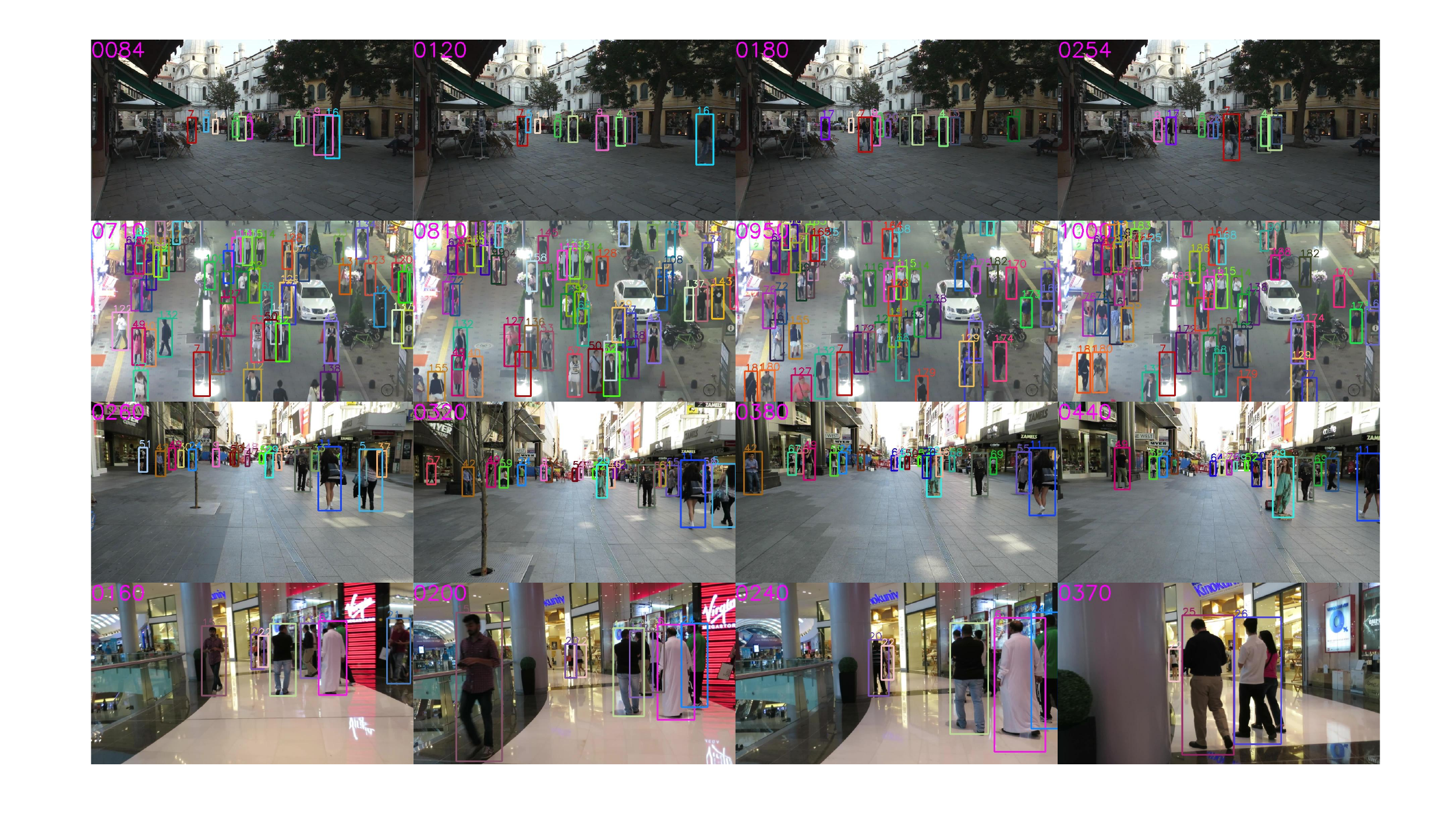}
\vspace{-1cm}
\end{center}
   \caption{Qualitative results of our tracking algorithm on MOT17 test split. Our method is able to track persons through long-term occlusions and also performs well in crowded scenes, best viewed in color.}
\label{fig:results}
\end{figure*}

On MOT17 benchmark, the work of DeepMOT~\cite{xu2020train} trains a neural network using the MOT Metric in an end-to-end manner. However their method is mainly optimized for improving two-frame data association. In contrast with their method, our method works in an offline manner which incorporates longer temporal contextual information both during training and inference. The results show that our method achieves better MOTA and IDF1 compared to those in~\cite{xu2020train}. A notable strong baseline is~\cite{papakis2020gcnnmatch}, which leverages a Graph Convolution Network followed by a Sinkhorn Network to perform end-to-end training of data association, we achieve similar performance in terms of MOTA compared to their approach but have better IDF1 and IDS score, due to the advantage of accurately learned costs as well as the multi-frame data association formulation. It is expected by further incorporating Graph Convolution Network (GCN) into feature learning, the performance of our method would be better. Figure~\ref{fig:results} shows some qualitative results, our method is able to track objects through long-term occlusions and recover missing detections.

Compared with the current SOTA method LifT~\cite{hornakova2020lifted}, our method achieves slightly worse MOTA and IDS metric. It should be noted that, the work of LifT considers a lifted connections between detections that span over 50 frames, making the formulation NP-hard and computationally heavy. By contrast, our method has polynomial time complexity. Although our method is slightly worse in-terms-of performance, we achieve faster inference speed: our tracker consumes 1-5 mins while their ILP solver requires 26.6 mins per sequence in average. We believe by working with a more powerful optimization technique such as multi-cut, multiple-hypothesis tracking during inference, our performance could be further leveraged.

Finally, we test our method on MOT20~\cite{dendorfer2020mot20}, which is designed for tracking the crowds. It is worth to mention that various pruning heuristics are applied in state-of-the-art method~\cite{hornakova2021making}, in order to make their NP-hard problem tractable, while our method does not require complex pre-processing steps to sparsify the graph. Overall, our method achieves second-best on MOTA and IDF1 metrics, which are slightly inferior than~\cite{hornakova2020lifted} but have better IDs metric than their results. Thanks to the multi-frame data association used during training/inference, our performance largely exceeds~\cite{bergmann2019tracking,papakis2020gcnnmatch,saleh2021probabilistic} in MOTA and MT metrics.

\vspace{-0.25cm}
\section{Conclusion}
\vspace{-0.1cm}
In summary, we have presented a general framework and a novel training method for learning the cost functions of the min-cost flow multi-object tracking problem. By solving a differentiable \textit{continuous} quadratic program (QP), our approach is able to incorporate multi-frame data association results as well as tracking specific constraints in order to obtain a better global objective for tracking. Although we perform tracking using network flow linear program inference, other formulations that respect tracking constraints can be employed for end-to-end learning as well. One major limitation of our method is that, only data association is learned end-to-end, whilst the part of object detection is separated from training. Since the success of this tracker largely relies on the quality of input detections, in the future we plan to explore training object detection jointly with our network flow framework.  We also aim to consider if the end-to-end learning of a higher order optimization objective can further improve the tracking performance.

\vspace{+0.9cm}
{\small
\bibliographystyle{ieee_fullname}
\bibliography{egbib}
}
\end{document}